\documentclass[letterpaper]{article} 
\usepackage{aaai25}  
\usepackage{times}  
\usepackage{helvet}  
\usepackage{courier}  
\usepackage[hyphens]{url}  
\usepackage{graphicx} 
\usepackage{natbib}  
\usepackage{caption} 
\usepackage{algorithm}
\usepackage{algorithmic}
\usepackage{booktabs}
\usepackage[table]{xcolor}
\usepackage{array}
\usepackage{amsmath,amssymb,amsfonts}
\usepackage{multirow}
\usepackage{newfloat}
\usepackage{listings}
\DeclareCaptionStyle{ruled}{labelfont=normalfont,labelsep=colon,strut=off} 
\floatstyle{ruled}
\newfloat{listing}{tb}{lst}{}
\floatname{listing}{Listing}
\title{Disentangling Tabular Data Towards Better One-Class Anomaly Detection}
\author{
    Jianan Ye\textsuperscript{\rm 1}$^{,}$\textsuperscript{\rm 2},
    Zhaorui Tan\textsuperscript{\rm 1}$^{,}$\textsuperscript{\rm 2},
    Yijie Hu\textsuperscript{\rm 1}$^{,}$\textsuperscript{\rm 2},
    Xi Yang\textsuperscript{\rm 1},
    Guangliang Cheng\textsuperscript{\rm 2},
    Kaizhu Huang\textsuperscript{\rm 3}\thanks{Corresponding author.}
}
\affiliations{
    \textsuperscript{\rm 1} School of Advanced Technology, Xi’an Jiaotong-Liverpool University\\
    \textsuperscript{\rm 2} School of Electrical Engineering, Electronics and Computer Science, University of Liverpool\\
    \textsuperscript{\rm 3} Data Science Research Center, Duke Kunshan University

    kaizhu.huang@dukekunshan.edu.cn
}

\usepackage{bibentry}

\begin{document}

\maketitle

\begin{abstract}
Tabular anomaly detection under the one-class classification setting poses a significant challenge, as it involves accurately conceptualizing ``normal" derived exclusively from a single category to discern anomalies from normal data variations.
Capturing the intrinsic correlation among attributes within normal samples presents one promising method for learning the concept. To do so, the most recent effort relies on a learnable mask strategy with a reconstruction task.
However, this wisdom may suffer from the risk of producing uniform masks, i.e., essentially nothing is masked, leading to less effective correlation learning.
To address this issue, we presume that attributes related to others in normal samples can be divided into two non-overlapping and correlated subsets, defined as CorrSets, to capture the intrinsic correlation effectively. Accordingly, we introduce an innovative method that disentangles CorrSets from normal tabular data.
To our knowledge, this is a pioneering effort to apply the concept of disentanglement for one-class anomaly detection on tabular data. Extensive experiments on 20 tabular datasets show that our method substantially outperforms the state-of-the-art methods and leads to an average performance improvement of 6.1\% on AUC-PR and 2.1\% on AUC-ROC.
Codes are available at \url{https://github.com/yjnanan/Disent-AD}.
\end{abstract}

\section{Introduction}

Tabular anomaly detection under the one-class classification setting presumes the availability of only one-class data, i.e., the normal class samples for training, while the goal during the test is to discern anomalies~\cite{chandola2009anomaly,ruff2021unifying}. In practical scenarios, such as financial fraud detection, cyber intrusion detection, and medical diagnosis~\cite{hilal2022financial,malaiya2019empirical,chen2022unsupervised}, this method tries to detect whether new data points conform to the pattern of observed normal samples, thereby identifying them as either normal or abnormal.
Given that only normal instances are available during the training, the inherent challenge lies in extracting the invariant feature of normal data. An instance observed with patterns deviating from these characteristics is detected as an anomaly. However, the lack of prior knowledge on structures in tabular data poses a significant challenge for learning such knowledge~\cite{shenkar2022anomaly}.

\begin{figure}[!t]
    \centering
    \includegraphics[width=\linewidth]{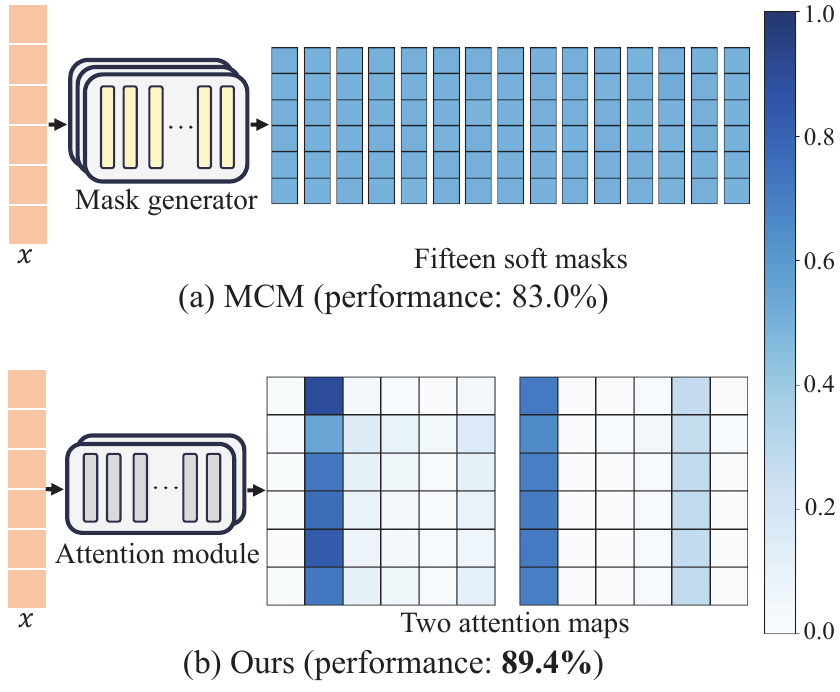}
    \caption{Visualization of MCM~\cite{yin2024mcm}'s fifteen soft masks and our two attention maps on the Thyroid dataset (six attributes). The presented masks and attention maps are averages derived from all training samples. In MCM's masks, darker colors indicate greater masking ratios while a higher attention weight in our attention maps.}
    \label{fig:banner}
\end{figure}

One reasonable way to mitigate this challenge is to focus on capturing the intrinsic correlation among attributes of normal samples.
Once one or more attributes of a sample deviate anomalously, it is an anomaly, and the correlation among attributes becomes different from that exhibited in normal samples. The most recent MCM~\cite{yin2024mcm} proposes a learnable mask generator module with a diversity loss to create diverse soft masks (i.e., mask values range from 0 to 1 for input data).
A reconstruction task is then designed to restore the original data from its masked variants.
During inference, anomalous attributes will disrupt the correlations observed in the normal class, leading to a failure to reconstruct the original data from masked instances, thereby allowing anomalies to be detected.
Despite its efficacy, the diversity loss may generate masks with nearly uniform values, inducing trivial solutions of the learned intrinsic correlations.
Empirical evidence in Figure~\ref{fig:banner}(a) suggests that the values of all soft masks are around $0.5$.
In this case, each attribute is scaled with the same value, indicating essentially no attribute is masked. Thus, the model learns the objective of reconstructing the original data from complete features instead of partial attributes, resulting in less effective extraction of the desired correlation.

In this paper, we presume that attributes related to others in normal tabular data can be split into two\footnote{It can be multiple subsets. However, our empirical results indicate using two subsets is the best.} non-overlapping and correlated subsets, termed as \textit{CorrSets}, to ensure the distillation of the intrinsic correlation among attributes from normal instances.
Building on this premise, this paper validates that disentangling those CorrSets from normal samples for reconstruction promotes the learning of normal samples' inside correlations.
Specifically, the sufficiently disentangled CorrSets are used to restore the whole original data individually, in which processes of the correlation inside normal data would be well captured by the model.
Compared to MCM~\cite{yin2024mcm}, which may generate ineffective masks, this approach guarantees that the model reconstructs the original data with partial attributes, thus strengthening its ability to extract the correlation within normal samples and yielding improvements in anomaly detection. 

Motivated by the aforementioned assumption, we propose a novel paradigm named Disent-AD, which efficiently disentangles CorrSets from normal tabular data and adequately
captures the intrinsic correlation.
For practical implementation, a two-head self-attention module is utilized to implicitly extract two distinct subsets in latent space.
The attention maps depict the attributes of interest to the network. By learning two independent attention maps, the two attention heads focus on different regions of tabular data, enabling the implicit extraction of distinct attribute subsets.
As shown in Figure~\ref{fig:banner}(b), darker regions in attention maps indicate greater weights. Evidently, the attention module successfully concentrates on two non-overlapping attribute subsets. To ensure that disentangled subsets are correlated, a reconstruction task is performed to restore the original data by utilizing subset features extracted from either of the two attention heads. Importantly, this reconstruction process is strategically employed to detect anomalies rather than aiming for a complete restoration of the original data.
During testing,  samples with high reconstruction errors are detected as anomalies.

Our contributions can be summarized as follows:
\begin{itemize}
    \item To our knowledge, this is one pioneering work that leverages the concept of disentanglement to enhance the efficiency of tabular one-class anomaly detection.
    \item We propose a novel paradigm that learns the intrinsic correlation by disentangling two distinct and correlated attribute subsets from normal tabular data.
    \item Extensive experiments conducted on 20 tabular datasets demonstrate the superiority of our method to state-of-the-art methods, with an average improvement of 6.1\% and 2.1\% in the AUC-PR and AUC-ROC, respectively.
\end{itemize}

\section{Related Work}
\textbf{One-class anomaly detection} aims to identify unobserved class samples during training by utilizing the knowledge learned from a single normal class.
Relying on distance measurement techniques~\cite{breunig2000lof}, empirical cumulative distribution functions~\cite{li2022ecod}, and regularized classifiers~\cite{scholkopf1999support} are some of the classical ways to tackle this task.
A straightforward approach involves modeling a distribution based on normal samples and evaluating the likelihood of each test sample~\cite{zong2018deep,li2020copod}.
Establishing a reliable decision boundary between normal and anomaly samples through an end-to-end process with an effective one-class loss function has proven useful~\cite{ruff2018deep}.
Generating pseudo anomalies is an alternative way to construct a decision boundary~\cite{goyal2020drocc,cai2022perturbation}, which utilizes adversarial training or variational autoencoder (VAE)~\cite{an2015variational} to produce synthetic data that are abnormal yet closely resemble normal samples.
Nevertheless, both methods require specific assumptions about distributions of normal and anomaly samples, which may not always be appropriate.

Recent methods focus on designing self-supervision learning tasks to detect anomalies, such as applying geometric transformations to images and predicting the transformation~\cite{golan2018deep}.
However, geometric transformations are essentially not suitable for tabular data.
To tackle this issue, GOAD~\cite{bergman2020classification} develops learnable transformations for tabular data.
Inspired by learnable transformations, NeuTraLAD~\cite{qiu2021neural} designs one contrastive loss to learn the invariant relationship among original and multiple transformed samples for normal data.
With the idea of building a criterion based on contrastive loss, ICL~\cite{shenkar2022anomaly} aims at maximizing the mutual information between each subset and the rest of the parts in normal tabular data.
SLAD~\cite{xu2023fascinating} proposes scale as a new characteristic for tabular data to capture the invariant representations of the normal class, which is treated as the relationship between the dimensionality of attribute subsets and that of their transformed latent representations.
The most recent MCM~\cite{yin2024mcm} focuses on learning the intrinsic correlation of normal samples by restoring original data from data masked by diverse soft masks.
Unlike previous methods, our work concentrates on implicitly disentangling CorrSets from normal data to extract the intrinsic correlation for anomaly detection.

\textbf{Semi- and weakly-supervised anomaly detection on tabular data} assumes that anomaly samples are accessible during training~\cite{yoon2020vime,chang2023data,pang2023deep}.
Introducing anomaly knowledge into the training process enables these methods to extract discriminative knowledge between normal and anomaly features.
In this paper, we train the model for anomaly detection without prior knowledge of anomalies.
This task is particularly challenging as it involves constructing a criterion for distinguishing between normal and abnormal samples without explicit examples of anomalies to guide the process.

\textbf{Out-of-distribution (OOD) detection} often trains the network with a conventional classification task (e.g., ten-class classification) and identifies test samples belonging to the different distribution from which the training set is sampled~\cite{liu2020energy,zhu2022boosting}.
Training with a multi-class classification task permits OOD detection methods to capture discriminative knowledge across various classes, thereby facilitating effective estimation of the training data distribution.
However, our task aims to train a network with only one class for anomaly detection, which is more challenging as learning from one class may lead the model to produce predictions arbitrarily.

\section{Methodology}
\subsection{Problem Statement}
One-class anomaly detection on tabular data setup involves a training set $\mathcal{D}_{train}=\{x_i \in {\mathbb{R}}^M\}_{i=1}^N$, which consists of $N$ normal samples with $M$ attributes. A test set,  $\mathcal{D}_{test}=\{x_i \in \mathbb{R}^M, y_i \in \mathcal{Y}\}_{i=1}^J$, consists of samples $x_i$ with labels $y_i$, where $\mathcal{Y}=\left\{0,1\right\}$ (0 denotes a normal and 1 denotes an anomaly).
By training on $\mathcal{D}_{train}$, a deep anomaly detection model constructs a scoring function $\phi : \mathbb{R}^M \rightarrow \mathbb{R}$ that quantitatively assesses the abnormality levels of new data points.

\subsection{Overview}
\begin{figure}[!t]
    \centering
    \includegraphics[width=\linewidth]{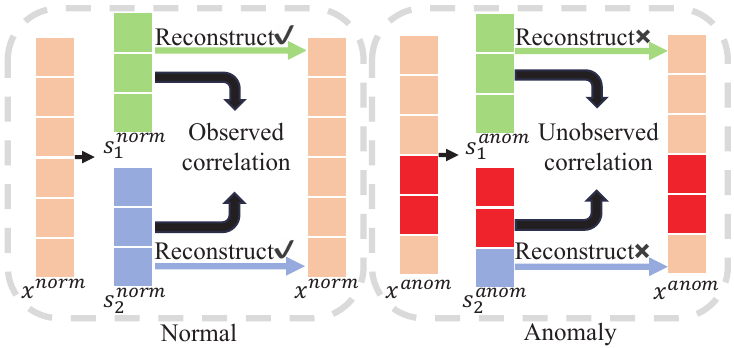}
    \caption{Illustration of our strategy. Red regions indicate anomaly attributes.}
    \label{fig:illustration}
\end{figure}

Our strategy for anomaly detection is illustrated in Figure~\ref{fig:illustration}.
Considering a tabular dataset where each sample is composed of $M$ attributes: $x_i = \{a_1, a_2, \cdots, a_M\}$. We denote the CorrSets of normal data as $s^{norm}_{i,1}$ and $s^{norm}_{i,2}$, $s^{norm}_{i,1} \cup s^{norm}_{i,2} = x_i^{norm}$ if there are no noise attributes in $x_i$.
The inherent correlation can be captured by training network $f_{\theta}$ to restore the original data with only $s^{norm}_1$ or $s^{norm}_2$, i.e. $f_{\theta}(s^{norm}_{i,1}) \rightarrow x_i^{norm}$ and $f_{\theta}(s^{norm}_{i,2}) \rightarrow x_i^{norm}$.
For anomaly instances during the test stage, we also extract two subsets, $s_1^{anom}$ and $s_2^{anom}$. If anomaly attributes are presented in $s_1^{anom}$ or $s_2^{anom}$, normal attributes of anomalies can not be reconstructed well, as the internal relations of anomalies deviate from those of normal instances.
There is a case where anomalous attributes are not in two subsets, but the model fails to restore them using normal attribute subsets, still producing large reconstruction errors.
Noisy attributes are unavoidable in real-world scenarios; however, neither normal nor anomaly attributes reconstruct noise well, which barely affects the distinction between normal and abnormal samples.
Accordingly, a criterion based on the reconstruction error for anomaly detection can be built by disentangling CorrSets from normal samples.
It is worth noting that our method leverages reconstruction to detect anomalies rather than fully restoring the original data.

\begin{figure}[!t]
    \centering
    \includegraphics[width=\linewidth]{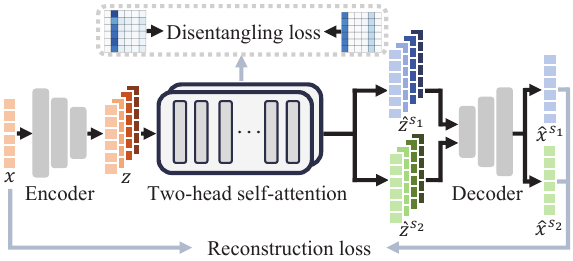}
    \caption{Illustration of our framework. The model consists of three modules: an encoder, a two-head self-attention, and a decoder. The network is trained under the constraint of a disentangling loss and a reconstruction loss.}
    \label{fig:framework}
\end{figure}

The framework of our method is presented in Figure~\ref{fig:framework}. Given one sample for illustrating our procedure, the encoder takes it as input and extracts its latent features. Then, a two-head self-attention module is utilized to disentangle tabular data for extracting features of two attribute subsets. Afterward, a decoder maps these features back to the input space for independently reconstructing the original data. Our model is trained in an end-to-end fashion and optimized by a disentangling loss and a reconstruction loss. During inference, the sum of the two subsets' reconstruction errors serves as the anomaly score for each test sample.

\subsection{Our Proposed Method}
To capture the correlation inside normal samples, we propose to disentangle CorrSets from them. There are two main components in our method: disentangling and learning correlation. Practicality, we leverage a two-head self-attention for implicit disentanglement, and a reconstruction task guarantees the correlation between two subsets.

\subsubsection{Disentangling}
Aiming to extract two distinct attribute subsets from normal samples, we utilize a two-head self-attention module to implicitly disentangle tabular data in latent space. The attention map describes attributes that the network focuses on, and the greater the attention weight, the more interested the network is. Learning two independent attention maps allows two heads of the attention module to extract latent features of two distinct attribute subsets in tabular data, thus enabling implicit disentanglement. We respectively demonstrate the importance of extracting two subsets and disentangling with two-head attention for capturing effective correlations in Section~\ref{sec:ablation}.

The attention module is applied to latent features extracted by the encoder to disentangle tabular data. To successfully adopt the attention module, we convert tabular data from row vectors into column vectors, i.e., $x_i \in \mathbb{R}^{M\times 1}$, where $M$ still denotes the number of attributes, and each attribute owns $1$ feature channel. The encoder $f_E$ maps $x_i$ to latent features $z_i \in \mathbb{R}^{M\times C}=f_E \left(x_i\right)$, where $C$ refers to the channel number of each attribute.
With the attention, $z_i$ yields queries, keys, and values, symbolized respectively as $q_i^{s_h}$, $k_i^{s_h}$, $v_i^{s_h}$ $\in \mathbb{R}^{M\times C}$, for each head $h \in \{1,2\}$. The computation of the two attention maps is formalized as follows:
\begin{equation}
    w_i^{s_h} = Softmax\left(\frac{q_i^{s_h}(k_i^{s_h})^T}{\sqrt{C}}\right), h \in \{1,2\},
\end{equation}
where the size of $w_i^{s_h}$ is $M\times M$. To accomplish the objective of disentangling two distinct attribute subsets from each training instance, it is imperative to learn two independent attention maps. Our strategy is to diminish the similarity between them. To this end, a cosine similarity loss is employed to yield a disentangling loss:
\begin{equation}
    \mathcal{L}_d = \frac{1}{N} \sum\nolimits_{i=1}^N \frac{w_i^{s_1}}{\|w_i^{s_1}\|_2} \cdot \frac{w_i^{s_2}}{\|w_i^{s_2}\|_2},
\end{equation}
where $\|\cdot\|_2$ is the $L_2$ norm. As $\mathcal{L}_d$ approaches 0, the two attention maps are orthogonal, thus allowing two heads to focus on distinct attribute subsets. Drawing from empirical observations, due to the absence of two attention maps with inverse relationships in single samples, the absolute value is not employed in the computation of $\mathcal{L}_d$. Consequently, this restriction confines the range of $\mathcal{L}_d$ to interval $[0,1]$, despite the range of the cosine similarity loss spanning $[-1,1]$. We validate the significance of $\mathcal{L}_d$ for learning the intrinsic correlation of normal data in Section~\ref{sec:ablation}.

\begin{figure}[!t]
    \centering
    \includegraphics[width=\linewidth]{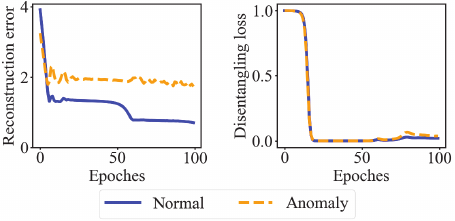}
    \caption{Averaged reconstruction error and disentangling loss of test samples during training on the Thyroid dataset.}
    \label{fig:score_vis}
\end{figure}

\subsubsection{Learning Correlation}
With disentangling, we extract two distinct attribute subsets from normal tabular data.
In order to capture the intrinsic correlation, it is essential to ensure that the disentangled subsets are correlated.
Therefore, we perform a reconstruction task to individually restore the input sample from the features of each subset.
Such a strategy guides the network in extracting CorrSets, thus progressing the learning of the intrinsic correlation of normal instances.

\begin{table*}[!t]
\centering
\small{
\begin{tabular}{@{}l|ccccccccccc>{\columncolor{gray!20}}c}
\toprule
Dataset          & IForest & LOF  & OCSVM         & ECOD          & DAGMM & DeepSVDD          & GOAD     & NeuTraLAD         & ICL               & SLAD              & MCM               & Ours              \\ \midrule
Arrhythmia       & 50.9    & 52.7 & 53.3          & 44.6          & 46.6  & 64.5$^{\pm4.1}$          & 63.2$^{\pm2.3}$ & 60.9$^{\pm4.5}$          & \textbf{65.4$^{\pm0.7}$} & 63.7$^{\pm1.5}$          & 58.3$^{\pm0.3}$          & 63.6$^{\pm0.3}$          \\
Breastw          & 94.4    & 99.2 & 99.3          & 95.2          & 75.8  & 97.0$^{\pm0.3}$         & 99.4$^{\pm0.1}$ & 85.3$^{\pm2.4}$          & 98.3$^{\pm0.5}$          & 98.9$^{\pm0.2}$          & 99.7$^{\pm0.0}$          & \textbf{99.8$^{\pm0.0}$} \\
Cardio           & 70.1    & 83.6 & \textbf{86.1} & 36.3          & 30.8  & 67.4$^{\pm5.4}$         & 31.4$^{\pm4.3}$ & 63.5$^{\pm2.5}$          & 59.5$^{\pm7.9}$          & 71.3$^{\pm1.9}$          & 78.5$^{\pm0.8}$          & 82.7$^{\pm2.8}$          \\
Census           & 13.5    & 23.4 & 22.7          & 17.7          & 10.6  & 18.8$^{\pm1.6}$          & 12.6$^{\pm0.6}$ & 13.6$^{\pm0.5}$          & 18.1$^{\pm0.3}$          & 13.8$^{\pm0.1}$          & 21.3$^{\pm0.8}$          & \textbf{39.7$^{\pm0.0}$} \\
Campaign         & 46.0    & 44.5 & 47.4          & 47.0          & 24.7  & 33.0$^{\pm3.3}$          & 47.1$^{\pm0.3}$ & 44.4$^{\pm2.7}$          & 43.5$^{\pm0.5}$          & 48.6$^{\pm0.0}$          & \textbf{59.0$^{\pm1.8}$} & 46.9$^{\pm1.0}$          \\
Cardiot. & 60.3    & 57.3 & 66.1          & 69.6          & 44.4  & 47.6$^{\pm5.5}$          & 68.6$^{\pm0.6}$ & 66.3$^{\pm0.9}$          & 65.5$^{\pm3.2}$          & 71.3$^{\pm3.8}$          & 69.7$^{\pm1.4}$          & \textbf{75.1$^{\pm0.9}$} \\
Fraud            & 69.3    & 40.4 & 34.9          & 40.6          & 0.9   & \textbf{73.5$^{\pm1.8}$} & 54.9$^{\pm0.7}$ & 58.2$^{\pm0.1}$          & 67.5$^{\pm0.9}$          & 54.3$^{\pm4.2}$          & 50.5$^{\pm2.2}$          & 60.5$^{\pm1.6}$          \\
Glass            & 9.5     & 9.2  & 8.9           & 11.1          & 10.1  & 29.3$^{\pm5.1}$          & 23.1$^{\pm3.3}$ & 24.0$^{\pm2.9}$          & 30.5$^{\pm2.0}$          & 22.2$^{\pm0.6}$          & 15.3$^{\pm0.6}$          & \textbf{58.9$^{\pm6.0}$} \\
Ionosphere       & 97.6    & 95.9 & 89.6          & 97.1          & 70.4  & 97.9$^{\pm0.4}$          & 93.0$^{\pm1.4}$ & \textbf{98.6$^{\pm0.4}$} & 96.9$^{\pm0.0}$          & \textbf{98.6$^{\pm0.3}$} & 97.7$^{\pm0.1}$          & 98.3$^{\pm0.2}$          \\
Mammo.      & 33.3    & 40.6 & 41.7          & \textbf{53.8} & 11.4  & 47.1$^{\pm2.7}$          & 22.7$^{\pm0.7}$ & 8.3$^{\pm0.5}$           & 19.5$^{\pm3.6}$          & 17.5$^{\pm3.9}$          & 47.9$^{\pm6.4}$          & 43.9$^{\pm1.5}$          \\
NSL-KDD          & 75.3    & 74.5 & 75.2          & 63.5          & 75.0  & 89.7$^{\pm4.3}$          & 85.7$^{\pm0.3}$ & 88.4$^{\pm2.4}$          & 60.4$^{\pm2.4}$          & \textbf{91.2$^{\pm0.5}$} & 90.2$^{\pm0.6}$          & 88.6$^{\pm0.8}$          \\
Optdigits        & 15.7    & 43.6 & 6.9           & 6.6           & 5.3   & 34.4$^{\pm9.5}$          & 8.1$^{\pm0.6}$  & 31.3$^{\pm14.9}$         & 85.3$^{\pm8.0}$          & 33.9$^{\pm2.9}$          & 80.2$^{\pm8.1}$          & \textbf{86.8$^{\pm4.8}$} \\
Pima             & 66.6    & 69.7 & 70.0          & 58.7          & 59.5  & 70.2$^{\pm1.2}$          & 63.7$^{\pm1.3}$ & 56.4$^{\pm1.4}$          & 60.0$^{\pm1.2}$          & 61.7$^{\pm0.7}$          & 69.4$^{\pm0.9}$          & \textbf{73.7$^{\pm2.7}$} \\
Pendigits        & 51.3    & 78.5 & 51.7          & 41.4          & 4.4   & 49.6$^{\pm8.1}$          & 50.0$^{\pm2.2}$ & 72.2$^{\pm6.8}$          & 52.4$^{\pm5.1}$          & 81.0$^{\pm4.7}$          & 74.2$^{\pm10.0}$         & \textbf{89.5$^{\pm6.6}$} \\
Satellite        & 85.8    & 80.8 & 77.7          & 83.3          & 68.6  & 80.7$^{\pm1.2}$          & 84.7$^{\pm0.1}$ & 87.3$^{\pm0.1}$          & 88.2$^{\pm0.5}$          & 87.5$^{\pm0.0}$          & 83.6$^{\pm0.0}$          & \textbf{89.1$^{\pm0.0}$} \\
Satimage-2       & 88.4    & 96.9 & 91.9          & 77.7          & 11.4  & 46.6$^{\pm35.1}$         & 96.3$^{\pm0.0}$ & 97.3$^{\pm0.3}$          & 96.8$^{\pm1.0}$          & 97.4$^{\pm0.3}$          & 97.8$^{\pm0.2}$          & \textbf{98.3$^{\pm0.1}$} \\
Shuttle          & 91.7    & 96.0 & 94.8          & 98.1          & 48.7  & 97.6$^{\pm0.6}$          & 91.6$^{\pm0.2}$ & \textbf{99.5$^{\pm0.2}$} & 97.3$^{\pm0.6}$          & 92.7$^{\pm0.5}$          & 97.8$^{\pm0.5}$          & \textbf{99.5$^{\pm0.2}$} \\
Thyroid          & 60.5    & 78.9 & 81.3          & 68.0          & 10.9  & 51.6$^{\pm2.4}$          & 49.2$^{\pm5.1}$ & 80.9$^{\pm0.3}$          & 83.6$^{\pm3.0}$          & 85.5$^{\pm1.6}$          & 83.0$^{\pm0.5}$          & \textbf{89.4$^{\pm0.7}$} \\
Wbc              & 85.7    & 84.1 & 83.9          & 72.1          & 29.5  & 72.9$^{\pm2.4}$          & 62.2$^{\pm7.9}$ & 29.1$^{\pm4.3}$          & 75.5$^{\pm5.5}$          & 46.3$^{\pm7.7}$          & 86.1$^{\pm1.2}$          & \textbf{91.0$^{\pm0.5}$} \\
Wine             & 24.5    & 12.5 & 14.2          & 35.7          & 49.0  & 55.0$^{\pm11.8}$         & 93.4$^{\pm1.6}$ & 85.7$^{\pm11.6}$         & 93.0$^{\pm4.6}$          & 99.6$^{\pm0.4}$          & 92.2$^{\pm0.4}$          & \textbf{99.9$^{\pm0.0}$} \\ \midrule
Average          & 59.5    & 63.1 & 59.9          & 55.9          & 34.4  & 61.2$^{\pm5.3}$          & 60.0$^{\pm1.6}$ & 62.5$^{\pm2.9}$          & 67.8$^{\pm2.5}$          & 66.8$^{\pm1.7}$          & 72.6$^{\pm1.8}$          & \textbf{78.7$^{\pm1.5}$}\\
Mean rank &7.6 &6.8 &7.0 &8.1 &11.2 &6.2 &7.3 &6.6 &5.9 &4.8 &4.0 &\textbf{2.0}\\
\bottomrule
\end{tabular}
}
\caption{Comparison of AUC-PR results (\%) of various  methods. The best result per dataset is \textbf{bold}. The mean rank ($\downarrow$) (bottom row) is calculated out of 12. Cardiot. and Mammo. refer to the Cardiotocography and Mammography datasets, respectively.}
\label{tab:aucpr}
\end{table*}

As a result of directing two attention heads to focus on distinct attribute subsets, the latent features of those can be obtained by:
\begin{equation}
    \hat{z}_i^{s_h} = w_i^{s_h} v_i^{s_h}, h\in \{1,2\},
\end{equation}
where $\hat{z}_i^{s_h} \in \mathbb{R}^{M\times C}$. Then a decoder $f_D$ takes $\hat{z}_i^{s_h}$ as inputs to produce reconstruction outputs:
\begin{equation}
    \hat{x}_i^{s_h} = f_D \left(\hat{z}_i^{s_h} \right), h\in \{1,2\}.
\end{equation}
The mean squared error (MSE) loss function is applied for the reconstruction loss:
\begin{equation}
    \mathcal{L}_r = \frac{1}{N} \sum\nolimits_{i=1}^N \sum\nolimits_{h=1,2} \left(x_i - \hat{x}_i^{s_h}\right)^2.
\end{equation}

The overall loss function for training our method is defined as follows:
\begin{equation}
    \mathcal{L}_{overall} = \mathcal{L}_d + \mathcal{L}_r,
\end{equation}
where we equal the weight for $\mathcal{L}_d$ and $\mathcal{L}_r$ to avoid a possible bias to one loss. By training with $\mathcal{L}_{overall}$, the model is able to extract CorrSets, thereby learning the inherent correlation of normal data.

\subsection{Anomaly Score for Inference}
The determination of a test sample as normal or abnormal is made by evaluating the reconstruction error.
The cosine similarity between attention maps is not included in the anomaly score function. It is due to that the model is trained to extract two distinct subsets. Empirical evidence is presented in Figure~\ref{fig:score_vis}. In the right part, the disentangling loss of anomalies varies almost identically to that of normal samples and eventually nears 0. As for the reconstruction error, the network aims to model the intrinsic correlation of normal class, thus test normal samples generally yield low errors. While anomalies consistently exhibit high reconstruction errors.
Therefore, we define the anomaly score function $\phi\left(x_i\right)$ with the MSE function as below:
\begin{equation}
    \phi\left(x_i\right) = \sum\nolimits_{h=1,2} \left(x_i - \hat{x}_i^{s_h}\right)^2.
\end{equation}
The anomaly scores for normal data are expected to be as small as possible.
For anomalies, whether or not the anomalous attributes are included in disentangled subsets, the model fails to reconstruct the complete input as the intrinsic correlation of the anomalies deviates from that of normal samples, resulting in high anomaly scores.
The greater the anomaly score, the more likely that a sample is an anomaly.

\section{Experiments}
\subsection{Datasets}
Our evaluation encompasses 20 tabular datasets, aligning with previous work~\cite{yin2024mcm}. 12 of them are obtained from the Outlier Detection Datasets (ODDS)~\cite{Rayana2016}, while the remainder are derived from ADBench~\cite{han2022adbench}, which cross various application scenarios, including healthcare, finance, and more. For specific dataset statistics, please see Supplementary D.

\subsection{Evaluation Metrics}
Experiments on tabular data are conducted by following previous work~\cite{xu2023fascinating, yin2024mcm}. We randomly sample 50\% of the normal samples as the training set, and the remaining normal samples with all anomaly samples are combined into the test set. Area Under the Precision-Recall Curve (AUC-PR) and Area Under the Receiver-Operating-Characteristic Curve (AUC-ROC) are selected as our evaluation criteria.
These two metrics can objectively evaluate detection performance without making any assumption on the decision threshold. All reported results are averaged over three independent trials.

\subsection{Implementation Details}
In our network architecture, a three-layer Multilayer Perceptron (MLP) with LeakyReLU activation function forms the encoder, and the decoder is symmetrically designed to the encoder. For the adaption of the attention module to tabular data, we transform row vectors into column vectors for most datasets and perform a patch-splitting preprocessing method for the rest to facilitate disentanglement, mostly derived from image scenarios. We consider the data structures of these datasets to be too complicated to affect disentangling. The preprocessing method is performed by splitting the data into three patches with the size of $M/2$ and treating each patch as an attribute to compose new data to simplify the data structure.
We present a detailed description of the preprocessing method in Supplementary A.
For datasets applied with the preprocessing method, we set epochs to 200 and the channel number $C$ of latent features to 512 for efficient convergence, while epochs to 100 and $C$ to 128 for the rest.
Due to the large variation in the number of samples between datasets, ranging from 129 to 299,285, we use different batch sizes for different datasets.
The parameters of the network are optimized by Adam with a uniform learning rate of 1e-4 for all datasets. See Supplementary B for more implementation details.

\begin{table*}[!t]
\centering
\small{
\begin{tabular}{@{}l|ccccccccccc>{\columncolor{gray!20}}c}
\toprule
Dataset                 & IForest       & LOF  & OCSVM         & ECOD & DAGMM & DeepSVDD & GOAD              & NeuTraLAD         & ICL      & SLAD              & MCM               & Ours               \\ \midrule
Arrhythmia       & 77.3          & 76.8 & 76.8          & 71.9 & 72.8  & 79.1$^{\pm3.0}$ & \textbf{81.3$^{\pm0.7}$} & 79.8$^{\pm0.9}$          & 81.0$^{\pm0.5}$ & 80.6$^{\pm0.3}$          & 78.9$^{\pm0.5}$          & 80.3$^{\pm0.2}$           \\
Breastw          & 97.1          & 99.3 & 99.3          & 96.4 & 76.0  & 97.4$^{\pm0.2}$ & 99.4$^{\pm0.1}$          & 90.8$^{\pm1.5}$          & 98.4$^{\pm0.3}$ & 98.9$^{\pm0.1}$          & \textbf{99.7$^{\pm0.0}$} & \textbf{99.7$^{\pm0.0}$}  \\
Cardio           & 92.2          & 95.6 & \textbf{96.5} & 63.7 & 69.5  & 85.4$^{\pm3.0}$ & 49.4$^{\pm3.8}$          & 85.9$^{\pm2.6}$          & 82.3$^{\pm4.7}$ & 87.2$^{\pm0.5}$          & 92.9$^{\pm0.6}$          & 95.3$^{\pm0.8}$           \\
Census           & 60.1          & 71.1 & 71.5          & 57.7 & 45.0  & 64.1$^{\pm1.4}$ & 58.0$^{\pm2.1}$          & 53.6$^{\pm1.0}$          & 67.1$^{\pm0.2}$ & 61.8$^{\pm0.4}$          & 73.6$^{\pm0.6}$          & \textbf{83.3$^{\pm0.0}$}  \\
Campaign         & 72.6          & 70.6 & 76.2          & 75.5 & 57.8  & 62.7$^{\pm2.2}$ & 73.0$^{\pm0.2}$          & 71.4$^{\pm4.1}$          & 70.7$^{\pm0.6}$ & 73.7$^{\pm0.1}$          & \textbf{88.0$^{\pm1.0}$} & 75.9$^{\pm1.7}$           \\
Cardiot. & 72.4          & 64.4 & 75.2          & 78.8 & 61.0  & 60.6$^{\pm5.8}$ & 81.3$^{\pm0.6}$          & 74.8$^{\pm1.6}$          & 70.3$^{\pm3.6}$ & 81.6$^{\pm3.6}$          & 81.1$^{\pm1.2}$          & \textbf{81.9$^{\pm0.2}$}  \\
Fraud            & \textbf{96.3} & 95.7 & 95.4          & 85.3 & 72.7  & 91.4$^{\pm2.0}$ & 93.8$^{\pm1.1}$          & 91.0$^{\pm0.9}$          & 92.6$^{\pm0.6}$ & 95.4$^{\pm0.2}$          & 94.0$^{\pm0.1}$          & 95.4$^{\pm0.2}$           \\
Glass            & 57.7          & 56.2 & 54.8          & 62.3 & 59.8  & 74.6$^{\pm3.2}$ & 75.5$^{\pm3.1}$          & 77.6$^{\pm3.1}$          & 86.2$^{\pm1.5}$ & 78.9$^{\pm2.5}$          & 72.1$^{\pm1.6}$          & \textbf{92.7$^{\pm0.3}$}  \\
Ionosphere       & 96.8          & 94.5 & 87.6          & 95.6 & 70.4  & 97.4$^{\pm0.5}$ & 91.8$^{\pm1.2}$          & \textbf{98.2$^{\pm0.4}$} & 96.5$^{\pm0.1}$ & \textbf{98.2$^{\pm0.3}$} & 96.7$^{\pm0.1}$          & 97.8$^{\pm0.2}$           \\
Mammo.      & 82.2          & 89.2 & 90.0          & 82.5 & 74.1  & 84.8$^{\pm0.9}$ & 73.5$^{\pm0.5}$          & 69.8$^{\pm1.5}$          & 57.9$^{\pm3.2}$ & 75.8$^{\pm2.3}$          & \textbf{91.2$^{\pm1.5}$} & 90.2$^{\pm0.3}$           \\
NSL-KDD          & 73.8          & 54.9 & 57.0          & 38.1 & 61.3  & 83.1$^{\pm8.4}$ & 84.6$^{\pm0.4}$          & 80.0$^{\pm5.7}$          & 31.5$^{\pm4.5}$ & 84.6$^{\pm0.8}$          & \textbf{87.9$^{\pm0.4}$} & 83.2$^{\pm0.5}$           \\
Optdigits        & 82.3          & 96.6 & 63.3          & 61.4 & 47.0  & 77.4$^{\pm4.3}$ & 68.0$^{\pm2.5}$          & 86.4$^{\pm5.7}$          & 99.0$^{\pm0.4}$ & 92.8$^{\pm0.2}$          & 98.9$^{\pm0.5}$          & \textbf{99.1$^{\pm0.2}$}  \\
Pima             & 67.3          & 69.1 & 71.3          & 58.3 & 61.0  & 63.9$^{\pm0.8}$ & 64.4$^{\pm1.1}$          & 55.2$^{\pm2.5}$          & 57.4$^{\pm1.7}$ & 61.5$^{\pm1.5}$          & 70.9$^{\pm1.5}$          & \textbf{75.8$^{\pm2.2}$}  \\
Pendigits        & 96.6          & 99.0 & 96.3          & 92.9 & 39.8  & 89.2$^{\pm3.0}$ & 95.4$^{\pm0.4}$          & 98.2$^{\pm0.9}$          & 91.9$^{\pm3.3}$ & 98.9$^{\pm0.4}$          & 98.9$^{\pm0.5}$          & \textbf{99.5$^{\pm0.3}$}  \\
Satellite        & 80.2          & 73.9 & 66.6          & 78.8 & 72.5  & 77.5$^{\pm1.2}$ & 78.0$^{\pm0.3}$          & 83.3$^{\pm0.2}$          & 85.5$^{\pm0.4}$ & 84.2$^{\pm0.6}$          & 78.7$^{\pm0.0}$          & \textbf{86.8$^{\pm0.1}$}  \\
Satimage-2       & 99.3          & 99.6 & 98.1          & 96.5 & 89.9  & 88.8$^{\pm9.4}$ & 99.6$^{\pm0.0}$          & 99.8$^{\pm0.0}$          & 99.7$^{\pm0.1}$ & 99.7$^{\pm0.1}$          & 99.8$^{\pm0.0}$          & \textbf{99.9$^{\pm0.0}$}  \\
Shuttle          & 99.6          & 99.8 & 99.6          & 99.7 & 90.4  & 99.2$^{\pm0.4}$ & 99.4$^{\pm0.0}$          & \textbf{99.9$^{\pm0.0}$} & 98.6$^{\pm0.4}$ & 99.5$^{\pm0.0}$          & \textbf{99.9$^{\pm0.0}$} & \textbf{99.9$^{\pm0.0}$}  \\
Thyroid          & 92.7          & 98.5 & 98.5          & 88.2 & 71.4  & 93.1$^{\pm1.2}$ & 93.4$^{\pm0.9}$          & 98.3$^{\pm0.1}$          & 98.4$^{\pm0.2}$ & 98.6$^{\pm0.3}$          & 97.8$^{\pm0.0}$          & \textbf{98.9$^{\pm0.0}$}  \\
Wbc              & 97.1          & 96.7 & 96.6          & 87.4 & 79.9  & 91.8$^{\pm1.6}$ & 89.4$^{\pm0.2}$          & 78.7$^{\pm1.4}$          & 91.5$^{\pm2.6}$ & 86.2$^{\pm2.6}$          & 97.3$^{\pm0.5}$          & \textbf{98.5$^{\pm0.1}$}  \\
Wine             & 65.7          & 40.8 & 48.5          & 74.3 & 88.3  & 84.4$^{\pm1.9}$ & 98.7$^{\pm0.4}$          & 96.8$^{\pm2.4}$          & 97.0$^{\pm3.0}$ & 99.9$^{\pm0.0}$          & 95.4$^{\pm0.9}$          & \textbf{100.0$^{\pm0.0}$} \\ \midrule
Average          & 83.0          & 82.1 & 81.0          & 77.3 & 68.0  & 82.2$^{\pm2.7}$ & 82.3$^{\pm0.9}$          & 83.4$^{\pm1.8}$          & 82.6$^{\pm1.5}$ & 86.9$^{\pm0.8}$          & 89.6$^{\pm0.5}$          & \textbf{91.7$^{\pm0.3}$}  \\ 
Mean rank &6.7 &6.2 &6.4 &8.8 &10.7 &8.0 &6.6 &6.7 &6.9 &4.5 &3.7 &\textbf{1.7} \\ 
\bottomrule
\end{tabular}}
\caption{AUC-ROC results of our method and competing methods.}
\label{tab:aucroc}
\end{table*}

\subsection{Baseline Methods}
Our method is compared with eleven outstanding tabular anomaly detection methods. IForest~\cite{liu2008isolation}, LOF~\cite{breunig2000lof}, OCSVM~\cite{scholkopf1999support}, and ECOD~\cite{li2022ecod} represent classical non-deep methods. The competing deep methods include DAGMM~\cite{zong2018deep}, DeepSVDD~\cite{ruff2018deep}, GOAD~\cite{bergman2020classification}, NeuTraLAD~\cite{qiu2021neural}, ICL~\cite{shenkar2022anomaly}, SLAD~\cite{xu2023fascinating}, and MCM~\cite{yin2024mcm}. It is noted that the results of IForest, LOF, OCSVM, ECOD, and DAGMM are obtained from MCM. We implement DeepSVDD, GOAD, NeuTraLAD, ICL, and SLAD by using DeepOD~\cite{xu2023deep}, an open-source Python library for deep learning-based anomaly detection. The implementation of MCM is based on their official open-source code, and original results reported in their paper are presented in Supplementary E.

\subsection{Anomaly Detection Performance}
Table~\ref{tab:aucpr} and~\ref{tab:aucroc} present the AUC-PR and AUC-ROC results of our method alongside the competing methods across 20 datasets, respectively. Despite the heterogeneity in the datasets, our method achieves the best overall performance on both two evaluation metrics, outperforming the second-best method by an average of 6.1\% on AUC-PR and 2.1\% on AUC-ROC. 
To prevent a few datasets from dominating the averaged results, we also present the mean rank for comparison. The best mean ranking out of 12 is obtained by our method, which is considerably lower than competing methods.
At the dataset level, our method beats competitors on 13 out of 20 datasets according to both AUC-PR and AUC-ROC, while exhibiting competitive performance on the remaining datasets. Significant performance gains are obtained by our method on some datasets compared to the second-best method, e.g., 16.3\% improvement in AUC-PR and 9.7\% in AUC-ROC on Census, and 28.4\% in AUC-PR and 6.5\% in AUC-ROC on Glass. Generally, these comparison results validate the superiority of our method.

\subsection{Ablation Study}
\label{sec:ablation}
Four datasets sourced from different application scenarios are selected to conduct the ablation study: Thyroid (healthcare), Arrhythmia (healthcare), Glass (forensic), and Satellite (satellite image). The results are reported in Table~\ref{tab:ablation}. 

\textbf{Disentangling two subsets is crucial for learning the intrinsic correlation of normal data:} We design ``Variant 1" that learns one attention map with a one-head self-attention module. Restoring original data from one subset makes the model extract attributes that can be used for reconstruction. However, the attention module might focus on each attribute to minimize reconstruction loss, causing the model to struggle with capturing the intrinsic correlation of normal samples. According to the experimental results, the detection performance of ``Variant 1" is much lower than that of our method, validating the significance of disentangling two subsets for learning the correlation in normal samples.

\begin{table}[!t]
\centering
\small{
\begin{tabular}{@{}cc|ccc>{\columncolor{gray!20}}c}
\toprule
\multicolumn{2}{c|}{Method}                                & Variant 1 & Variant 2 & Variant 3 & Ours     \\ \midrule
\multicolumn{2}{c|}{Two subsets}                           & -         & \checkmark & \checkmark & \checkmark         \\
\multicolumn{2}{c|}{Two heads}                             & -         & -          & \checkmark & \checkmark        \\
\multicolumn{2}{c|}{$\mathcal{L}_d$}                       & -         & -          &  -         & \checkmark         \\ \midrule
\multicolumn{1}{c|}{\multirow{4}{*}{\rotatebox{90}{AUC-PR}}}  & Thyroid    & 72.6$^{\pm1.6}$  & 78.8$^{\pm4.0}$  & 72.4$^{\pm1.2}$  & \textbf{89.4$^{\pm0.7}$} \\
\multicolumn{1}{c|}{}                         & Arrhythmia & 59.4$^{\pm1.7}$  & 61.7$^{\pm0.8}$  & 58.7$^{\pm0.9}$  & \textbf{63.6$^{\pm0.3}$} \\
\multicolumn{1}{c|}{}                         & Glass      & 40.8$^{\pm16.1}$ & 44.2$^{\pm7.0}$  & 31.3$^{\pm8.9}$  & \textbf{58.9$^{\pm6.0}$} \\
\multicolumn{1}{c|}{}                         & Satellite  & 88.4$^{\pm0.0}$  & 88.6$^{\pm0.3}$  & 87.5$^{\pm0.2}$  & \textbf{89.1$^{\pm0.0}$} \\ \midrule
\multicolumn{1}{c|}{\multirow{4}{*}{\rotatebox{90}{AUC-ROC}}} & Thyroid    & 97.1$^{\pm0.1}$  & 95.9$^{\pm1.3}$  & 97.3$^{\pm0.1}$  & \textbf{98.9$^{\pm0.0}$} \\
\multicolumn{1}{c|}{}                         & Arrhythmia & 75.4$^{\pm1.7}$  & 79.1$^{\pm0.5}$  & 75.0$^{\pm2.0}$  & \textbf{80.3$^{\pm0.2}$} \\
\multicolumn{1}{c|}{}                         & Glass      & 80.3$^{\pm15.7}$ & 88.4$^{\pm3.3}$  & 76.3$^{\pm6.8}$  & \textbf{92.7$^{\pm0.3}$} \\
\multicolumn{1}{c|}{}                         & Satellite  & 85.7$^{\pm0.0}$  & 86.5$^{\pm0.3}$  & 85.3$^{\pm0.2}$  & \textbf{86.8$^{\pm0.1}$} \\ \bottomrule
\end{tabular}}
\caption{Ablation study of our method and its variants.}
\label{tab:ablation}
\end{table}

\textbf{Extracting two subsets with two-head self-attention can capture better correlation than with one-head self-attention:} ``Variant 2" disentangles two subsets with a one-head self-attention module. We implement it by $w_i^{s_2} = 1 - w_i^{s_1}$, where $w_i^{s_1}$ is produced by the attention module, and they are independent of each other. Evidently, ``Variant 2" is significantly inferior to our method. It is due to that noise attributes might be contained in subsets according to $w_i^2$, leading to less effective learning of the intrinsic correlation. In our method, disentangling with the two-head self-attention module helps reduce the effect of noise attributes on learning the correlation. Additionally, the performance gap between ``Variant 2" and ``Variant 1" further demonstrates the significance of disentangling two subsets.

\textbf{Constraining the two subsets to be distinct is necessary for learning the correlation within normal data:} A substantial performance drop is observed in ``Variant 3", since the absence of the disentangling loss hinders the model from extracting two non-overlapping attribute subsets. This indicates that the disentangling loss is crucial in our method. We report the results of replacing our disentangling loss with MCM~\cite{yin2024mcm}’s diversity loss in Supplementary E.

\begin{figure}[!t]
    \centering
    \includegraphics[width=\linewidth]{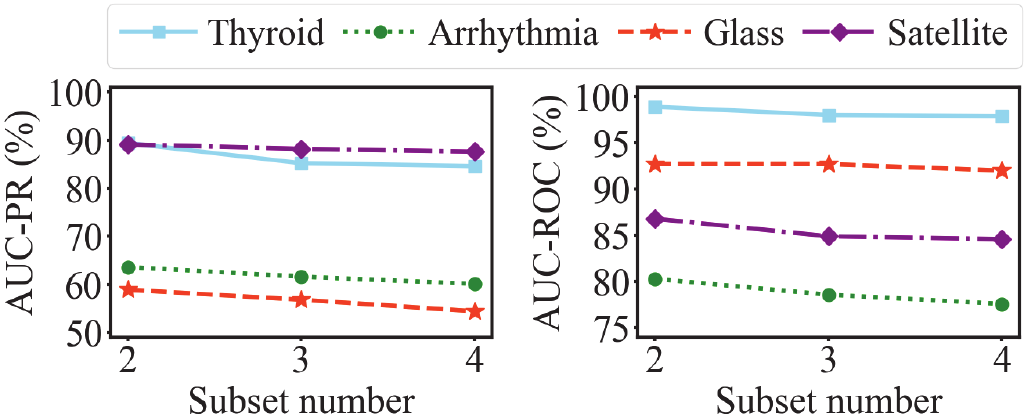}
    \caption{AUC-PR and AUC-ROC \textit{vs.}  subset numbers.}
    \label{fig:subsetnumber}
\end{figure}

\subsection{Further Analysis}
\subsubsection{Analysis on Subset Number}
Figure~\ref{fig:subsetnumber} reports the detection performance of our method vs. different numbers of disentangled attribute subsets, where Thyroid, Arrhythmia, Glass, and Satellite are used as illustrative datasets. Evidently, as the subset number is increased, the performance drops slightly, partially as extracting more correlated subsets raises the difficulty of training convergence. Additionally, there is a case that multiple ($>2$) subsets related to each other are not present in normal samples, which may lead to computational redundancy. This analysis indicates the rationale of our two subsets strategy. Visualization of attention maps for three subsets is shown in Supplementary C.

\begin{figure}[!t]
    \centering
    \includegraphics[width=\linewidth]{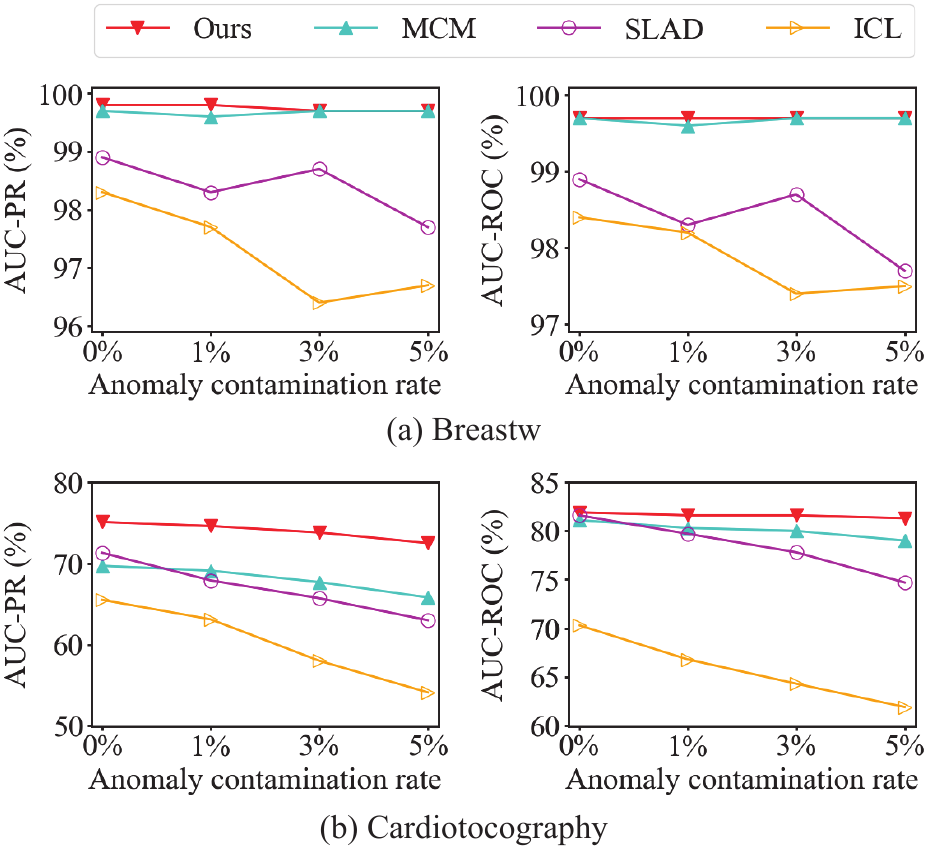}
    \caption{AUC-PR and AUC-ROC \textit{vs.} various ratios of anomaly contamination.}
    \label{fig:robustness}
\end{figure}

\subsubsection{Robustness to Anomaly Contamination}
In real-world anomaly detection, the training set may be contaminated, with a small percentage of anomalies included. By reference to~\cite{yin2024mcm}, we successively set the anomaly contamination ratio to 0\%, 1\%, 3\%, and 5\% to analyze the robustness of our method to anomaly contamination. Selecting Breastw and Cardiotocography as typical datasets, analysis results are shown in Figure~\ref{fig:robustness}. It is observed that our method experiences minimal performance degradation as the anomaly contamination ratio increases, consistently yielding the best results in each case. Such empirical evidence justifies the outstanding robustness of our method to anomaly contamination.

\begin{table}[!t]
\centering
\small{
\begin{tabular}{c|cc}
\toprule
Data shape & Satellite         & Satimage-2          \\ \midrule
$M\times1^*$       & 75.7$^{\pm0.2}$/74.6$^{\pm0.5}$ & 96.5$^{\pm0.2}$/99.3$^{\pm0.1}$ \\
$2\times M/2$       & 87.7$^{\pm0.5}$/84.5$^{\pm0.0}$ & 98.1$^{\pm0.1}$/\textbf{99.9$^{\pm0.0}$} \\
$2\times 3M/4$       & 85.9$^{\pm0.1}$/81.7$^{\pm0.3}$ & 98.2$^{\pm0.1}$/\textbf{99.9$^{\pm0.0}$} \\
$3\times M/3$       & 86.7$^{\pm0.3}$/82.5$^{\pm0.6}$ & 97.3$^{\pm0.1}$/99.7$^{\pm0.1}$ \\
\rowcolor{gray!20}
$3\times M/2$ (Ours)       & \textbf{89.1$^{\pm0.0}$}/86.8$^{\pm0.1}$ & \textbf{98.3$^{\pm0.1}$}/\textbf{99.9$^{\pm0.0}$} \\ \midrule
Shuffling  & 88.5$^{\pm0.5}$/\textbf{86.9$^{\pm0.3}$} & 97.9$^{\pm0.0}$/99.8$^{\pm0.0}$ \\ \bottomrule
\end{tabular}}
\caption{AUC-PR/AUC-ROC results on Satellite and Satimage datasets with varying data shapes after applying the preprocessing method. $M$ denotes the attributes number in the original data, and * denotes results without preprocessing. Shuffling refers to randomly shuffling the order of attributes in the original data before using preprocessing.}
\label{tab:preprocess}
\end{table}

\subsubsection{Analysis on the Preprocessing Method}
Table~\ref{tab:preprocess} presents the detection performance of preprocessing data into various data shapes, where Satellite and Satimage-2 are selected as illustrative datasets. They are sourced from satellite image scenarios and preprocessed in our implementation. There are overlaps between split patches when the data shape is $2\times 3M/4$ or $3\times M/2$. Evidently, there is a noticeable performance improvement when applying the preprocessing method, as a reduced number of attributes enables the model to disentangle data more easily, thereby facilitating correlation learning. According to the empirical results, we use the data shape of $3 \times M/2$ in our implementation.
We also test our method's robustness to randomly ordered attributes in original data when using the preprocessing method. Although there is a slight performance drop, the results are still comparable, which verifies that our method is robust to the randomly ordered attributes.

\section{Conclusion}
In this paper, we design a novel disentangling-based method to learn the correlation inside normal tabular data by extracting two distinct and correlated attribute subsets for anomaly detection. To our knowledge, this is the first work to successfully leverage disentanglement for tabular anomaly detection under the one-class classification setting. Extensive experiments conducted on 20 tabular datasets sourced from diverse application scenarios with AUC-PR and AUC-ROC metrics evidence that our method outperforms the existing best methods.
However, it is imperative to note that our current design specifically caters to tabular data and, as such, does not directly apply to other data types, such as image or point cloud data. We intend to explore disentangling-based methods for non-tabular data types in future work.

\section*{Acknowledgments}
The work was partially supported by the following: National Natural Science Foundation of China under No. 92370119, No. 62376113, and No. 62206225; The Alan Turing Institute (UK) through the project 'Turing-DSO Labs Singapore Collaboration' (SDCfP2\textbackslash100009).

\bibliography{aaai25}

\newpage

\appendix

\section{Patch-splitting Preprocessing}
\label{supp:preprocess}
\begin{figure}[!ht]
    \centering
    \includegraphics[width=\linewidth]{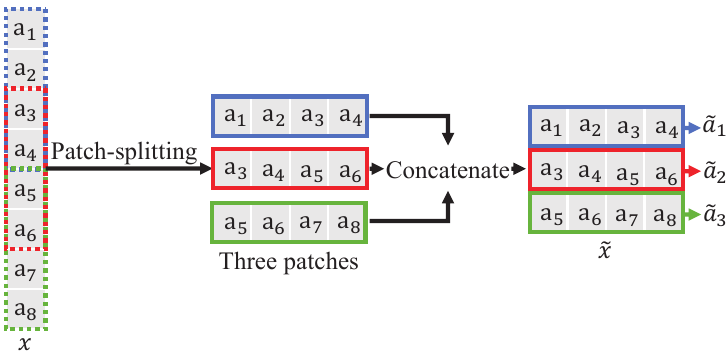}
    \caption{Illustration of the patch-splitting preprocessing.}
    \label{fig:preprocess}
\end{figure}

The patch-splitting preprocessing is performed to facilitate disentanglement, as illustrated in Figure~\ref{fig:preprocess}. In our implementation, we split the original data into three patches, each of size $M/2$, where $M$ is the attribute number of the original data. These patches are then concatenated to form new data, with each patch treated as a new attribute of the preprocessed data. With such a procedure, the original data is transformed into new data with fewer attributes and more features of each attribute, allowing the model to disentangle data more easily.

\section{More Implementation Details}
\label{supp:detail}
\subsection{Experimental Setup}
We conduct experiments on a single NVIDIA GeForce RTX 3090 GPU on the Ubuntu 20.04.1 system. Our code is implemented based on PyTorch 1.13.0 framework with Python 3.9.17. Other critical package requirements include numpy 1.25.0, scikit-learn 1.2.2, scipy 1.10.1, and pandas 2.0.3.

\subsection{Hyperparameters for Each Dataset}
\label{supp:hyperparameters}
Learning rate, batch size, training epoch, channel number $C$ of latent features, and preprocessed or not for different tabular datasets are presented in Table~\ref{tab:hyperparameters}. Datasets marked with \checkmark are applied with the patch-splitting preprocessing method to produce data with the shape of $3\times M/2$. We set epochs to 200 and $C$ to 512 for these datasets to ensure efficient convergence. The batch size of each dataset is set according to its number of training samples. The greater the training set, the larger the batch size.

\subsection{Analysis on Epoch and $C$}
Table~\ref{tab:epochC} presents the results with various epochs and $C$, where Satellite and Satimage-2 are illustrative datasets. Evidently, using the larger epoch and $C$ can obtain performance improvement, which validates the effectiveness of our hyperparemeter setting for epoch and $C$.

\subsection{Analysis on Batch size}
We report results with various batch sizes for each dataset in Table~\ref{tab:bsz}, which demonstrates the rationale of our batch size hyperparameter setting. Additionally, our method surpasses all competing methods when using a fixed batch size across all datasets.
It is noted that our method performs better with a smaller batch size on the Fraud dataset. However, to use a uniform batch size for datasets with a large number of training samples, we set the batch size to 2,048 for the Fraud dataset in our implementation.

\section{Attention Maps for Three Subsets}
\label{supp:3subsets}
\begin{figure}[!ht]
    \centering
    \includegraphics[width=\linewidth]{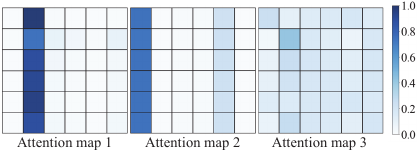}
    \caption{Visualization of three attention maps when disentangling three subsets from normal data on the Thyroid dataset.}
    \label{fig:3subsets}
\end{figure}

We visualize the average attention maps on training data of the Thyroid dataset when three subsets are disentangled from normal data in Figure~\ref{fig:3subsets}. Nearly uniform weights are observed in the third attention map, i.e., the third attention head focuses on complete data rather than on partial attributes, resulting in less effective correlation learning. This empirical evidence supports the rationale of our two subsets strategy.

\section{Statistics of Datasets}
\label{supp:statistic}
Statistics of 20 tabular datasets are presented in Table~\ref{tab:statistics}.

\section{More Results}
MCM's original results reported in their paper are presented in Table~\ref{tab:mcm}. Although MCM's results are slightly higher than our reproduced results, our method still substantially outperforms it.

Table~\ref{tab:dl} reports AUC-PR/AUC-ROC results of replacing our disentangling loss with MCM's diversity loss on Thyroid and Satellite datasets. Evidently, our method still outperforms MCM, demonstrating the superiority of our disentangling strategy.

\begin{table}[!ht]
\centering
\small{
\begin{tabular}{@{}l|ccccc@{}}
\toprule
Dataset          & Lr & Epoch & Bs & $C$   & Preprocessing \\ \midrule
Arrhythmia       & 1e-4        & 100    & 64         & 128 & -             \\
Breastw          & 1e-4        & 100    & 64         & 128 & -             \\
Cardio           & 1e-4        & 100    & 128        & 128 & -             \\
Census           & 1e-4        & 100    & 2,048       & 128 & -             \\
Campaign         & 1e-4        & 100    & 2,048       & 128 & -             \\
Cardiotocography & 1e-4        & 100    & 128        & 128 & -             \\
Fraud            & 1e-4        & 200    & 2,048       & 512 & \checkmark              \\
Glass            & 1e-4        & 100    & 64         & 128 & -             \\
Ionosphere       & 1e-4        & 200    & 64         & 512 & \checkmark              \\
Mammography      & 1e-4        & 100    & 2,048       & 128 & -             \\
NSL-KDD          & 1e-4        & 100    & 2,048       & 128 & -             \\
Optdigits        & 1e-4        & 200    & 128        & 512 & \checkmark              \\
Pima             & 1e-4        & 100    & 128        & 128 & -             \\
Pendigits        & 1e-4        & 200    & 128        & 512 & \checkmark              \\
Satellite        & 1e-4        & 200    & 512        & 512 & \checkmark              \\
Satimage         & 1e-4        & 200    & 512        & 512 & \checkmark              \\
Shuttle          & 1e-4        & 200    & 2,048       & 512 & \checkmark              \\
Thyroid          & 1e-4        & 100    & 512        & 128 & -             \\
Wbc              & 1e-4        & 100    & 64         & 128 & -             \\
Wine             & 1e-4        & 100    & 64         & 128 & -             \\ \bottomrule
\end{tabular}}
\caption{Hyperparameters for each dataset. Lr and Bs denote the learning rate and batch size, respectively. $C$ refers to the channel number of latent features.}
\label{tab:hyperparameters}
\end{table}

\begin{table}[!ht]
\centering
\small{
\begin{tabular}{@{}l|ccc@{}}
\toprule
Dataset          & $N$      & $D$   & Anomalies \\ \midrule
Arrhythmia       & 452    & 274 & 66 (14.6\%)       \\
Breastw          & 683    & 9   & 239 (34.9\%)      \\
Cardio           & 1,831   & 21  & 176 (9.6\%)      \\
Census           & 299,285 & 500 & 18,568 (6.2\%)    \\
Campaign         & 41,188  & 62  & 4,640 (11.2\%)     \\
Cardiotocography & 2,114   & 21  & 466 (22.0\%)      \\
Fraud            & 284,807 & 29  & 492 (0.1\%)      \\
Glass            & 214    & 9   & 9 (4.2\%)        \\
Ionosphere       & 351    & 33  & 126 (35.8\%)      \\
Mammography      & 11,183  & 6   & 260 (2.3\%)      \\
NSL-KDD          & 148,517 & 122 & 77,054 (51.8\%)    \\
Optdigits        & 5,216   & 64  & 150 (2.8\%)      \\
Pima             & 768    & 8   & 268 (34.8\%)      \\
Pendigits        & 6,870   & 16  & 156 (2.2\%)      \\
Satellite        & 6,435   & 36  & 2,036 (31.6\%)     \\
Satimage-2       & 5,803   & 36  & 71 (1.2\%)       \\
Shuttle          & 49,097  & 9   & 3,511 (7.1\%)     \\
Thyroid          & 3,772   & 6   & 93 (2.4\%)       \\
Wbc              & 278    & 30  & 21 (7.5\%)       \\
Wine             & 129    & 13  & 10 (7.7\%)       \\ \bottomrule
\end{tabular}}
\caption{Statistics of 20 tabular datasets. $N$ represents the total number of samples, $D$ denotes the number of attributes in each dataset, and the last column specifies the number of anomaly samples within each dataset.}
\label{tab:statistics}
\end{table}

\begin{table}[!ht]
\centering
\small{
\begin{tabular}{c|c|c}
\toprule
(Epoch, $C$) & Satellite         & Satimage-2        \\ \midrule
(100, 128)  & 88.3$^{\pm0.1}$/85.2$^{\pm0.3}$ & 97.0$^{\pm0.1}$/99.1$^{\pm0.0}$ \\
(100, 512)  & 89.0$^{\pm0.2}$/86.4$^{\pm0.4}$ & 97.1$^{\pm0.0}$/99.1$^{\pm0.0}$ \\
(200, 128)  & 89.0$^{\pm0.0}$/86.2$^{\pm0.1}$ & 97.3$^{\pm0.4}$/99.3$^{\pm0.3}$ \\
\rowcolor{gray!20}
(200, 512)  & \textbf{89.1$^{\pm0.0}$}/\textbf{86.8$^{\pm0.1}$} & \textbf{98.3$^{\pm0.1}$}/\textbf{99.9$^{\pm0.0}$} \\ \bottomrule
\end{tabular}}
\caption{AUC-PR/AUC-ROC results on Satellite and Satimage-2 datasets with varying epochs and $C$.}
\label{tab:epochC}
\end{table}

\begin{table}[!ht]
\centering
\small{
\begin{tabular}{@{}l|ccc@{}}
\toprule
Dataset          & MCM$^\dag$                & MCM$^*$       & Ours               \\ \midrule
Arrhythmia       & 58.3$^{\pm0.3}$/78.9$^{\pm0.5}$  & 61.0/81.1 & 63.6$^{\pm0.3}$/80.3$^{\pm0.2}$  \\
Breastw          & 99.7$^{\pm0.0}$/99.7$^{\pm0.0}$  & 99.5/99.5 & 99.8$^{\pm0.0}$/99.7$^{\pm0.0}$  \\
Cardio           & 78.5$^{\pm0.8}$/92.9$^{\pm0.6}$  & 84.8/96.0 & 82.7$^{\pm2.8}$/95.3$^{\pm0.8}$  \\
Census           & 21.3$^{\pm0.8}$/73.6$^{\pm0.6}$  & 24.2/75.8 & 39.7$^{\pm0.0}$/83.3$^{\pm0.0}$  \\
Campaign         & 59.0$^{\pm1.8}$/88.0$^{\pm1.0}$  & 60.4/89.0 & 46.9$^{\pm1.0}$/75.9$^{\pm1.7}$  \\
Cardiot. & 69.7$^{\pm1.4}$/81.1$^{\pm1.2}$  & 69.9/80.0 & 75.1$^{\pm0.9}$/81.9$^{\pm0.2}$  \\
Fraud            & 50.5$^{\pm2.2}$/94.0$^{\pm0.1}$  & 51.4/93.5 & 60.5$^{\pm1.6}$/95.4$^{\pm0.2}$  \\
Glass            & 15.3$^{\pm0.6}$/72.1$^{\pm1.6}$  & 19.0/72.2 & 58.9$^{\pm6.0}$/92.7$^{\pm0.3}$  \\
Ionosphere       & 97.7$^{\pm0.1}$/96.7$^{\pm0.1}$  & 98.0/97.2 & 98.3$^{\pm0.2}$/97.8$^{\pm0.2}$  \\
Mammo.      & 47.9$^{\pm6.4}$/91.2$^{\pm1.5}$  & 47.5/90.5 & 43.9$^{\pm1.5}$/90.2$^{\pm0.3}$  \\
NSL-KDD          & 90.2$^{\pm0.6}$/87.9$^{\pm0.4}$  & 90.8/87.8 & 88.6$^{\pm0.8}$/83.2$^{\pm0.5}$  \\
Optdigits        & 80.2$^{\pm8.1}$/98.9$^{\pm0.5}$  & 88.8/99.4 & 86.8$^{\pm4.8}$/99.1$^{\pm0.2}$  \\
Pima             & 69.4$^{\pm0.9}$/70.9$^{\pm1.5}$  & 73.8/76.3 & 73.7$^{\pm2.7}$/75.8$^{\pm2.2}$  \\
Pendigits        & 74.2$^{\pm10.0}$/98.9$^{\pm0.5}$ & 82.5/99.1 & 89.5$^{\pm6.6}$/99.5$^{\pm0.3}$  \\
Satellite        & 83.6$^{\pm0.0}$/78.7$^{\pm0.0}$  & 85.3/79.6 & 89.1$^{\pm0.0}$/86.8$^{\pm0.1}$  \\
Satimage-2       & 97.8$^{\pm0.2}$/99.8$^{\pm0.0}$  & 98.5/99.9 & 98.3$^{\pm0.1}$/99.9$^{\pm0.0}$  \\
Shuttle          & 97.8$^{\pm0.5}$/99.9$^{\pm0.0}$  & 94.7/99.7 & 99.5$^{\pm0.2}$/99.9$^{\pm0.0}$  \\
Thyroid          & 83.0$^{\pm0.5}$/97.8$^{\pm0.0}$  & 84.1/98.0 & 89.4$^{\pm0.7}$/98.9$^{\pm0.0}$  \\
Wbc              & 86.1$^{\pm1.2}$/97.3$^{\pm0.5}$  & 88.8/98.1 & 91.0$^{\pm0.5}$/98.5$^{\pm0.1}$  \\
Wine             & 92.2$^{\pm0.4}$/95.4$^{\pm0.9}$  & 93.3/95.3 & 99.9$^{\pm0.0}$/100.0$^{\pm0.0}$ \\ \midrule
Average          & 72.6$^{\pm1.8}$/89.6$^{\pm0.5}$  & 74.8/90.4 & 78.7$^{\pm1.5}$/91.7$^{\pm0.3}$  \\ \bottomrule
\end{tabular}}
\caption{AUC-PR/AUC-ROC results of MCM and our method, \dag refers to our reproduced results, * denotes original results reported in their paper.}
\label{tab:mcm}
\end{table}

\begin{table}[!ht]
\centering
\small{
\begin{tabular}{c|cc}
\toprule
Method                & Thyroid            & Satellite          \\ \midrule
MCM                   & 83.0/97.8          & 83.6/78.7          \\
Ours (diversity loss) & 88.3/98.6          & 88.9/86.2          \\
\rowcolor{gray!20}
Ours                  & \textbf{89.4/98.9} & \textbf{89.1/86.8} \\ \bottomrule
\end{tabular}}
\caption{AUC-PR/AUC-ROC results of replacing our disentangling loss with MCM’s diversity loss.}
\label{tab:dl}
\end{table}

\begin{table*}[!ht]
\centering
\small{
\begin{tabular}{@{}l|cccc@{}}
\toprule
Batch size       & 64                 & 128                & 512               & 2,048              \\ \midrule
Arrhythmia       & 63.6$^{\pm0.3}$/80.3$^{\pm0.2}$  & 63.3$^{\pm0.7}$/79.4$^{\pm0.7}$  & 63.6$^{\pm1.8}$/79.0$^{\pm0.2}$ & -                 \\
Breastw          & 99.8$^{\pm0.0}$/99.7$^{\pm0.0}$  & 99.6$^{\pm0.1}$/99.5$^{\pm0.1}$  & 99.6$^{\pm0.1}$/99.6$^{\pm0.1}$ & -                 \\
Cardio           & 81.0$^{\pm1.2}$/94.8$^{\pm0.2}$  & 82.7$^{\pm2.8}$/95.3$^{\pm0.8}$  & 80.6$^{\pm3.4}$/95.0$^{\pm0.0}$ & 78.8$^{\pm0.9}$/95.0$^{\pm0.5}$ \\
Census           & 30.1$^{\pm7.8}$/79.4$^{\pm4.2}$  & 36.4$^{\pm4.5}$/82.7$^{\pm0.6}$  & 39.5$^{\pm0.0}$/83.2$^{\pm0.0}$ & 39.7$^{\pm0.0}$/83.3$^{\pm0.0}$ \\
Campaign         & 44.1$^{\pm0.4}$/74.3$^{\pm0.7}$  & 44.7$^{\pm0.6}$/74.9$^{\pm2.7}$  & 45.7$^{\pm0.9}$/75.6$^{\pm1.8}$ & 46.9$^{\pm1.0}$/75.9$^{\pm1.7}$ \\
Cardiotocography & 75.0$^{\pm1.7}$/81.8$^{\pm2.6}$  & 75.1$^{\pm0.9}$/81.9$^{\pm0.2}$  & 68.4$^{\pm1.3}$/68.9$^{\pm4.1}$ & 68.0$^{\pm0.1}$/66.4$^{\pm0.3}$ \\
Fraud            & 69.5$^{\pm0.8}$/96.9$^{\pm0.3}$  & 67.4$^{\pm1.4}$/96.4$^{\pm0.1}$  & 63.5$^{\pm2.7}$/95.8$^{\pm0.1}$ & 60.5$^{\pm1.6}$/95.4$^{\pm0.2}$ \\
Glass            & 58.9$^{\pm6.0}$/92.7$^{\pm0.3}$  & 56.8$^{\pm5.9}$/91.6$^{\pm5.9}$  & -                 & -                 \\
Ionosphere       & 98.3$^{\pm0.2}$/97.8$^{\pm0.2}$  & 98.1$^{\pm0.2}$/97.4$^{\pm0.1}$  & -                 & -                 \\
Mammography      & 40.3$^{\pm1.3}$/84.0$^{\pm4.2}$  & 41.2$^{\pm4.9}$/85.0$^{\pm1.6}$  & 42.9$^{\pm2.9}$/85.3$^{\pm4.1}$ & 43.9$^{\pm1.5}$/90.2$^{\pm0.3}$ \\
NSL-KDD          & 83.4$^{\pm0.8}$/76.7$^{\pm3.6}$  & 86.5$^{\pm1.9}$/80.1$^{\pm4.9}$  & 87.5$^{\pm0.6}$/82.7$^{\pm0.9}$ & 88.6$^{\pm0.8}$/83.2$^{\pm0.5}$ \\
Optdigits        & 84.2$^{\pm7.3}$/99.0$^{\pm0.4}$  & 86.8$^{\pm4.8}$/99.1$^{\pm0.2}$  & 84.9$^{\pm8.7}$/98.9$^{\pm0.2}$ & 37.9$^{\pm5.3}$/91.3$^{\pm1.7}$ \\
Pima             & 73.3$^{\pm1.3}$/74.4$^{\pm1.2}$  & 73.7$^{\pm2.7}$/75.8$^{\pm2.2}$  & 72.6$^{\pm0.3}$/74.2$^{\pm0.3}$ & -                 \\
Pendigits        & 88.3$^{\pm1.2}$/99.2$^{\pm0.2}$  & 89.5$^{\pm6.6}$/99.5$^{\pm0.3}$  & 86.8$^{\pm0.8}$/99.2$^{\pm0.0}$ & 83.7$^{\pm1.7}$/98.9$^{\pm0.0}$ \\
Satellite        & 88.8$^{\pm0.3}$/86.0$^{\pm0.6}$  & 88.9$^{\pm0.3}$/86.6$^{\pm0.3}$  & 89.1$^{\pm0.0}$/86.8$^{\pm0.1}$ & 88.7$^{\pm0.3}$/85.9$^{\pm0.4}$ \\
Satimage         & 97.1$^{\pm0.2}$/99.4$^{\pm0.1}$  & 97.3$^{\pm0.4}$/99.4$^{\pm0.3}$  & 98.3$^{\pm0.1}$/99.9$^{\pm0.0}$ & 97.0$^{\pm0.0}$/99.2$^{\pm0.0}$ \\
Shuttle          & 99.5$^{\pm0.2}$/99.8$^{\pm0.1}$  & 99.5$^{\pm0.1}$/99.8$^{\pm0.0}$  & 99.5$^{\pm0.2}$/99.9$^{\pm0.0}$ & 99.5$^{\pm0.2}$/99.9$^{\pm0.0}$ \\
Thyroid          & 87.8$^{\pm1.1}$/98.6$^{\pm0.1}$  & 88.0$^{\pm0.6}$/98.6$^{\pm0.1}$  & 89.4$^{\pm0.7}$/98.9$^{\pm0.0}$ & 89.2$^{\pm0.8}$/98.9$^{\pm0.0}$ \\
Wbc              & 91.0$^{\pm0.5}$/98.5$^{\pm0.1}$  & 90.3$^{\pm0.7}$/98.4$^{\pm0.0}$  & -                 & -                 \\
Wine             & 99.9$^{\pm0.0}$/100.0$^{\pm0.0}$ & 99.9$^{\pm0.0}$/100.0$^{\pm0.0}$ & -                 & -                 \\
\toprule
Average          & 77.6$^{\pm1.6}$/90.6$^{\pm0.9}$  &78.2$^{\pm2.0}$/91.0$^{\pm1.0}$   &77.8$^{\pm1.5}$/90.5$^{\pm0.8}$  & 75.1$^{\pm1.1}$/90.1$^{\pm0.6}$ \\
\bottomrule
\end{tabular}}
\caption{AUC-PR/AUC-ROC results on each dataset with varying batch sizes. Results for some datasets with batch sizes of 512 or 2,048 are not shown, as the batch size is much larger than the number of training samples. For example, the Arrhythmia training set has 193 samples. As such, the entire training set are used in a single iteration when the batch size is 512 or 2,048. The results in these two cases are essentially the same.}
\label{tab:bsz}
\end{table*}

\end{document}